\documentclass[%
 aip,
 amsmath,amssymb,
 reprint,%
]{revtex4-1}

\usepackage{graphicx}
\usepackage{dcolumn}
\usepackage{bm}

\usepackage[utf8]{inputenc}
\usepackage[T1]{fontenc}
\usepackage{mathptmx}
\usepackage{xcolor}
\usepackage{comment}
\usepackage{multirow}
\usepackage{makecell}

\newcommand{\new}[1]{\textcolor{black}{#1}}

\begin{document}


\title{Materials representation and transfer learning for multi-property prediction}

\author{Shufeng Kong}
\affiliation{ Department of Computer Science, Cornell University, Ithaca, NY, USA}
\author{Dan Guevarra}
\affiliation{ Division of Engineering and Applied Science, California Institute of Technology, Pasadena, CA, USA}
\author{Carla P. Gomes}
\email{gomes@cs.cornell.edu}
\affiliation{ Department of Computer Science, Cornell University, Ithaca, NY, USA}
\author{John M. Gregoire}
\email{gregoire@caltech.edu}
\affiliation{ Division of Engineering and Applied Science, California Institute of Technology, Pasadena, CA, USA}


\begin{abstract}
The adoption of machine learning in materials science has rapidly transformed materials property prediction. Hurdles limiting full capitalization of recent advancements in machine learning include
the limited development of methods to learn the underlying interactions of multiple elements, as well as the relationships among multiple properties, to facilitate property prediction in new composition spaces. To address these issues, we introduce the Hierarchical Correlation Learning for Multi-property Prediction (H-CLMP) framework that seamlessly integrates (i) prediction using only a material’s composition, (ii) learning and exploitation of correlations among target properties in multi-target regression, and (iii) leveraging training data from tangential domains via generative transfer learning. The model is demonstrated for prediction of spectral optical absorption of complex metal oxides spanning 69 3-cation metal oxide composition spaces. H-CLMP accurately predicts non-linear composition-property relationships in composition spaces for which no training data is available, which broadens the purview of machine learning to the discovery of materials with exceptional properties. This achievement results from the principled integration of latent embedding learning, property correlation learning, generative transfer learning, and attention models. The best performance is obtained using H-CLMP with Transfer learning (H-CLMP(T)) wherein a generative adversarial network is trained on computational density of states data and deployed in the target domain to augment prediction of optical absorption from composition. H-CLMP(T) 
aggregates multiple knowledge sources with a framework that is well-suited for multi-target regression across the physical sciences.

\end{abstract}

\maketitle

\section*{Significance statement}

Machine learning algorithms have largely been developed to automate tasks that are routine for humans, which stands in contrast to scientific discovery where one seeks to predict something that has never been observed. For example, in materials science the chemistry of new combinations of 3 elements can yield properties beyond those observed by combining any 2 of the elements. We present a model based on hierarchical correlation learning that makes \new{some} such discoveries by learning chemical interactions from other domains via transfer learning. \new{The work aims to emulate}
a scientist's aggregation of multiple knowledge sources to 
\new{make informed predictions in new spaces.}

\section*{Introduction}

Recent progress in \new{the} development of artificial intelligence (AI), especially deep learning techniques, is broadening the purview of AI to address critical challenges in a broad range of fields.\cite{bohannon_cyberscientist_2017,gil_amplify_2014,lecun_deep_2015,schrittwieser_mastering_2020, decost_scientific_2020} The challenges of AI for materials science are extensive, making acceleration of materials discovery a formidable task that requires advancing the frontier of AI.\cite{choudhury_role_2020,liu_machine_2020,saal_machine_2020,tian_efficient_2021,nikolaev_autonomy_2016,alberi_2019_2018,kusne_--fly_2020,chen_learning_2021} Materials discovery embodies the convergence of limited data, data dispersed over multiple domains, and multi-property prediction, motivating commensurate integration of AI methods to collectively address these challenges, as demonstrated by the Hierarchical Correlation Learning for Multi-property Prediction (H-CLMP, pronounced ``H-Clamp'') framework introduced herein.

In molecular materials, structure is central to the definition of the material, and the lack of structural periodicity has enabled development of representations and associated machine learning models that are quite adept at property prediction.\cite{krenn_self-referencing_2020,kearnes_molecular_2016,qiao_orbnet_2020,chen_graph_2019} These concepts are increasingly being adapted for solid state materials, often involving materials representations that combine properties of the constituent elements with structural features.\cite{seko_representation_2017,ward_including_2017,schutt_schnet_2018, xie_crystal_2018, park_developing_2020, lym_lattice_2019, noh_inverse_2019,zhao_predicting_2020,ward_matminer_2018,louis_graph_2020} While these approaches are being deployed to good effect, they are not applicable in the common scenario of experimental materials science wherein a given material is composed of a mixture of phases, or even more so when no knowledge of the phases is available. In exploratory research for materials with specific properties, measurement and interpretation of composition and property data is often far less expensive than measurement and interpretation of structural data. Consequently, the need to accelerate exploratory research motivates development of models that derive their materials representation from only composition. 

Ward et al.\cite{ward_general-purpose_2016} developed the Magpie featurizer to create a high-dimensional vector representation of an any-dimensional composition by combining known elemental properties using a suite of manually-chosen functions, which \new{facilitated prediction of glass formability to guide}
experimental investigation in new ternary metal composition spaces.\cite{ren_accelerated_2018} This approach creates a deterministic vector for each composition that can be generically used for prediction models, although a machine-learned representation of composition using data relevant to the target prediction task is preferable.

Provided sufficient labelled training data, models such as deep neural networks (DNNs) can be effectively deployed. In this case, formulating explicit representations for composition may be unnecessary because the model can internalize representation learning. ElemNet embraces this approach and is an effective model in specific settings,\cite{jha_elemnet_2018} but more generally, the available training data for materials property prediction is typically insufficient to employ such brute force machine learning approaches.
Therefore, there is a need to create high-performance learners trained with more easily-obtained data from different but related domains. This methodology is referred to as transfer learning.\cite{zhuang_comprehensive_2021} Transfer learning aims to improve the performance of target learners on target domains by transferring the knowledge
contained in different but related source domains.
ElemNet was recently expanded to employ a type of transfer learning wherein model training commences with computational data and continues with final training in an experimental domain, where the target properties are common to the transfer-source and target-prediction domains.\cite{jha_enhancing_2019} Goodall et al. adopted a distinct approach named Roost wherein a vector representation of each element can be learned in the transfer-source domain to initialize a graph representation of composition that is further trained in the desired prediction domain.\cite{goodall_predicting_2020}
\new{While the Roost model concatenates the element embedding vectors with the element composition vector as an input to the attention component, a similar approach named CrabNet 
\new{adds}
a fractional encoding of the element composition vector  
to the element embedding vectors as an input to the attention component. \cite{wang2020compositionally}} In the present work, we employ transfer learning by training a generative adversarial network (GAN) \new{in a domain that is distinct from the primary prediction task, yet this ``transfer domain'' contains knowledge that may be relevant to the prediction task. The resulting model encodes composition-property relationships in the transfer domain and is used in the prediction domain to augment the input of the prediction model.}
This generative transfer learning enables utilization of any source domain with composition and property data by transferring the generative model itself and deploying it as an on-demand generator of the source-domain properties' representation in the target domain.

The machine learning models for materials property prediction reported to-date largely focus on single-property regression, wherein a scalar value is predicted by a given model. The most common domain is prediction of formation energy,\cite{ward_general-purpose_2016} or minimum formation energy when considering only composition,\cite{goodall_predicting_2020} which can accelerate search for new thermodynamically-stable phases.
More generally, prediction of multiple properties is desired, and since the underlying chemistry of the specific mixture of elements dictates the relationship between the input material representation and a given property, it stands to reason that multi-property prediction may be effectively achieved with frameworks that learn, harness, and exploit relationships among multiple properties. We tackle this challenge with correlation learning, \new{which has been demonstrated to enhance multi-label classification and multi-target regression.\cite{chen2018end,bai2020disentangled,zhao2021hot,kong2020deep} Since} the multiple properties being predicted may not be explicitly correlated, we developed a framework to learn correlations in latent embeddings of the multiple properties. 

Our primary setting involves prediction of multiple properties in a composition space
for which there is no available training data. Each of the elements in the prediction space appears in the training data, but the specific combination of the elements was never seen by the model during training. Human experts use various expressions of chemical similarity of the elements to hypothesize the properties of new combinations of elements, motivating development of machine learning models that learn the relationships among elements that are most pertinent to the properties being predicted. In 
machine learning,
attention mechanisms are used to
learn the interactions and relative importance of features for
a given task. Soft-attention builds
upon this concept by allowing the function that produces the
attention coefficients to be learned directly from the \new{data.\cite{gonzalez2021,lanchantin2019neural} }
\new{Roost and CrabNet use graph attention networks (GAT) wherein the nodes are elements, enabling learning of interactions among elements. H-CLMP(T) additionally uses GAT where the nodes are multi-property embeddings to learn relationships among multiple materials properties. }

The recent proliferation of deep learning for materials property prediction demonstrates the rapid advancement  of AI for materials discovery,\cite{schutt_schnet_2018, xie_crystal_2018, park_developing_2020, lym_lattice_2019, noh_inverse_2019,zhao_predicting_2020,jha_enhancing_2019,goodall_predicting_2020} and the present work makes a leap forward through seamless integration of machine learning techniques in a framework specifically designed to address the challenges of materials discovery. Our combination of attention learning, learning correlations among multiple predicted properties, and \new{generative} transfer learning enables multi-property prediction in new compositions spaces with better accuracy than state-of-the-art methods. The H-CLMP model, and its transfer-learning extension, H-CLMP(T), are introduced in the present work and demonstrated for discovery of materials with unique optical properties. The multiple properties of the prediction task are the measured optical absorption coefficients for a series of 10 photon energies that span the visible spectrum and extend into the ultraviolet. The materials whose properties are being predicted are complex metal oxides containing 3 cation elements (in addition to oxygen). For the high throughput synthesis and optical characterization methods that generated the training dataset, the cation composition and optical characterization comprise the only existing knowledge of the materials, requiring optical absorption prediction using only cation composition. 
 
The high throughput experiments that produced the dataset for the present work have been used to demonstrate neural network-based autoencoders for optical properties,\cite{stein_machine_2018} although the prediction  of optical absorption from composition for such data had yet to be demonstrated. This task is emblematic of materials multi-property prediction and encapsulates the quintessential goal for AI-accelerated materials discovery, to predict the properties of never-before-seen materials. \new{Progress toward this goal is demonstrated through evaluation of prediction accuracy for 69 data instances with available ground truth, each comprising a 3-cation composition space that is absent from the training data. Given the predictive power of H-CLMP(T) in this setting, we additionally deploy the model to predict spectral absorption in  129060 new 3-cation metal oxide compositions for which no experimental data is available. With additional discussion of use cases for these predictions, our introduction of H-CLMP(T) for predicting spectral absorption demonstrates that appropriately-constructed machine learning models can guide materials discovery in new composition spaces per the vision of AI-accelerated materials discovery.}

\section*{Results}

The presentation of results commences with a description of the H-CLMP model and introduction of the settings for which it is deployed, followed by analysis of multi-target regression performance compared to baseline models.

\subsection*{Hierarchical Correlation Learning for Multi-Property
Prediction}

\begin{figure*}
\centering
\includegraphics[width=0.85\linewidth]{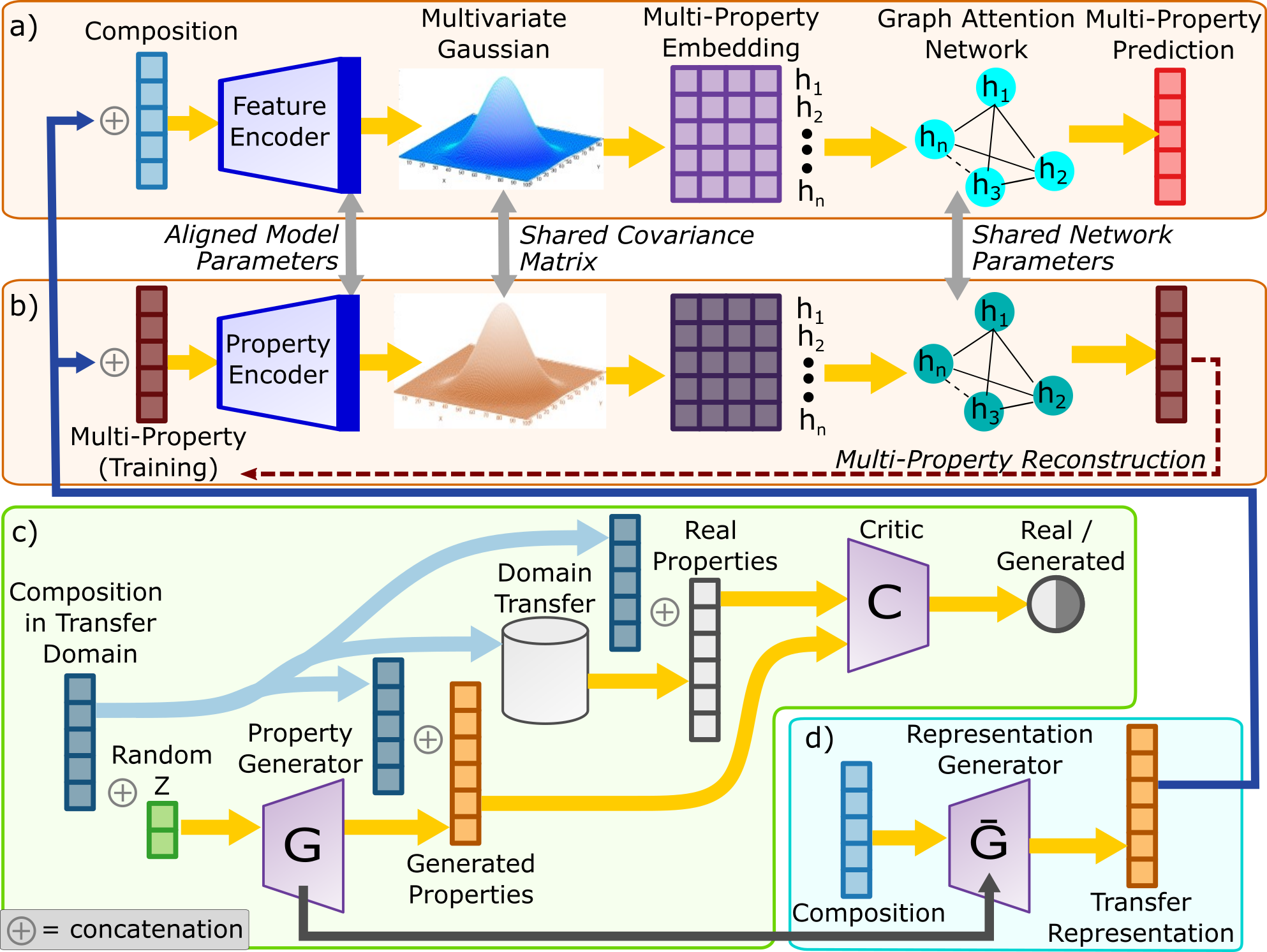}
\caption{
The H-CLMP(T) framework. Components (a) and (b) are jointly-trained parallel models for multi-property prediction and  multi-property reconstruction, 
respectively, \new{where component (b) is a variational autoencoder}. 
\new{The latent representations produced by the encoders are aligned during model training. }
\new{Each decoder commences with a multivariate Gaussian model for learning pairwise property correlations, where the covariance matrix is shared between components (a) and (b).}
Higher-order multi-property correlations are learned from the multi-property embeddings via \new{a graph attention network (GAT)}. Component (a) performs the desired multi-property prediction task, while the multi-property reconstruction of component (b) facilitates training \new{of component (a).}
Training and deployment of transfer learning is achieved by components (c) and (d), respectively. In (c), a conditional WGAN is trained with transfer domain data and used to construct a transfer representation generator in (d), which augments the model inputs in (a) and (b). Collectively, these components comprise H-CLMP(T) and are implemented by first training model (c) and then using (d) while jointly training (a) and (b). Deployment of the trained model to make new predictions proceeds by evaluating (d) and then (a) for a given composition.
}
\label{fig:H-CLMP}
\end{figure*}

\begin{table*}
\begin{tabular}{|c|c|c|c|c|c|c|c|}
\hline
 & \makecell{Feature \\ encoding} & Decoder & \makecell{Pairwise property \\ correlation learning} & \makecell{Higher-order property \\ correlation learning} & \makecell{Model \\ alignment} & \makecell{Transfer \\ learning}   & Ref.\\\hline
H-CLMP(T) & GAT & \makecell{Multivariate \\ Gaussian + GAT}  & \makecell{Multivariate \\ Gaussian} & GAT & VAE & \makecell{generative \\ via cWGAN} & \makecell{this \\ work} \\ \hline
Roost & GAT & ResNet  & NA & NA & NA & warm-up & \cite{goodall_predicting_2020}\\ \hline
CrabNet & \makecell{GAT with \\ fractional encoding} & ResNet & NA & NA & NA & warm-up & \cite{wang2020compositionally} \\ \hline
ElemNet & fractional encoding & MLP & NA & NA & NA & warm-up & \cite{jha_enhancing_2019} \\ \hline
\end{tabular}
\caption{\label{tab:models-compare} \new{The H-CLMP(T) model of the present work is compared to reported models for composition-to-property prediction using 6 model attributes. The decoders of prior models use standard ResNet or MLP networks, whereas H-CLMP(T) combines a multivariate Gaussian with a graph attention network (GAT) for correlation learning, both of which are trained via alignment with a multi-property variational autoencoder (VAE). H-CLMP(T) uses a conditional Wasserstein generative adversarial network (cWGAN) to transfer learned embeddings from another domain, whereas previous methods have used other domains to warm-up the models, i.e. pre-train model parameters.
} 
}
\end{table*}

Multi-property prediction (MPP) aims to predict the values of multiple targets, and while independent single-property prediction models can be used for each target, this approach is both inefficient and unable to harness relationships among the functional mappings from input to properties. Harnessing and exploiting this knowledge requires the model to incorporate techniques such as representation and correlation learning.
Exploring the correlations among properties and learning rich property and feature embeddings are two important aspects of MPP as they  
can yield
a more accurate predictor. We developed a machine learning \new{framework} to address these two important aspects of MPP, which involves materials representation learning and transfer learning. The overall architecture of our model for MPP from materials composition is presented in Figure~\ref{fig:H-CLMP}, which includes four highly interdependent components: (a) the multi-property prediction model, with composition and transferred data as input, jointly trained  with (b) an aligned target property autoencoder;
(c) a separately trained generator for transfer learning; and (d) the deployment of the generator for transfer learning. \new{Components (a) and (b) comprise the core H-CLMP model, with the addition of (c) and (d) comprising it's transfer-learning extension, H-CLMP(T). Transfer learning is a term generally used when information from one domain is brought to bear on the target prediction task. In the present work, we consider only the most aggressive type of transfer learning, where both the input materials and output properties of the transfer domain are distinct from the target domain.}

The MPP model integrates several machine learning techniques to tackle specific challenges. The model was designed to seamlessly integrate 2 primary hierarchies of correlation learning, commencing with a multivariate Gaussian \new{whose} covariance matrix $\Sigma$  \new{is} shared across all (composition, properties) data points. The use of individual $\mu$ parameters for each composition but a global $\Sigma$ compels the multivariate Gaussian model to generate embeddings that capture relationships among the functional mappings from composition to properties. This shared covariance matrix is a compact, \new{interpretable} latent space that can be learned with modest quantities of training data. The multi-property embeddings generated by the multivariate Gaussian model serve as input to another correlation learning module, a \new{GAT}. This type of model excels at learning local and high-order correlations, complementing the previous \new{pairwise} correlation learning.

To facilitate training of (a), especially in limited data regimes, model training is performed in conjunction with component (b), 
a variational auto-encoder (VAE), \new{which further conditions the model}. The VAE learns a latent space that is used 
\new{to regularize the latent space of (a), which is achieved by adding a Kullback–Leibler (KL) loss between the latent representations of (a) and (b) to the training loss function.}
As a result, the latent space learned by component (a) implicitly encodes the element compositions and material properties. Also, sharing the network parameters between the attention graph neural networks in components (a) and (b) mitigates 
issues with
local optima during model training. 

Components (a) and (b) \new{ are design for }
MPP  within a single domain \new{where the model inputs can be any available representation of the materials. In the target domain, the metal oxide cation composition is the only available representation of the material, and components (c) and (d) employ generative transfer learning to derive a more expressive representation of each material in the target domain.} 
The central model of component (c) is $G$, a generative adversarial network (GAN) trained exclusively in the transfer domain, which in the present case is computational density of states data from the Materials Project (MP-DOS).\cite{jain_materials_2013} Specifically, model $G$ is a conditional Wasserstein GAN (cWGAN)
that learns the conditional distribution of properties given composition. In the present use case, $G$ encodes the relationships among elements pertaining to the MP-DOS. 

\new{If computational data for spectral absorption were available, it could be used for components (c) and (d), which would still be an aggressive transfer learning setting wherein computational spectral absorption on periodic structures is used to represent complex metal oxide compositions for predicting experimental spectral absorption. We are not aware of a computational spectral absorption dataset with substantial breadth in metal oxide composition, which is understandable given the computational expense of calculating the probabilities for optically-driven transitions in the joint density of states.\cite{naccarato_searching_2019} As a result, we are using MP-DOS, for which no explicit relationship between these properties and those of the target domain is known. However, our materials chemistry intuition is that if combining certain cations provides similar/different features in the computational DOS, this combination of cations will be more/less likely to have similar experimental spectral absorption compared to a random combination of cation elements. This intuition is the basis for using the cWGAN to learn conditional probability distributions, which is not necessarily the optimal strategy for accurately predicting MP-DOS in component (c) but is an effective means of capturing elemental interactions in model $G$, which encompasses the knowledge that is transferred to the target domain to facilitate the MPP task. As an aside, the cWGAN of component (c) provides a standardized mean absolute error of 0.43 on a random 10\% holdout of the MP-DOS entries, demonstrating that the cWGAN has substantial predictive power in this domain and motivating further exploration of this approach for DOS prediction in future work.}

The transfer representation generator $\bar{G}$ of component (d) is fixed after training of component (c). There are many complementary strategies for the implementation of $\bar{G}$ given the trained model $G$. Herein, the transfer representation (output of $\bar{G}$) is the average property vector generated by sampling $G$'s distribution. 
For each composition used in components (a) and (b), component (d) generates a representation of the transfer domain knowledge, which is concatenated to the inputs of H-CLMP to complete the generative transfer learning. \new{As a result, in the transfer learning setting, each material in the target domain is represented by the concatenation of composition as well as its generated MP-DOS from component (d).}
   
For the MPP task, H-CLMP(T) is compared to \new{baseline models based on 3 different approaches for property prediction from composition, which are compared to H-CLMP in TABLE~\ref{tab:models-compare}}.
\new{One such baseline model is} ElemNet,\cite{jha_elemnet_2018,jha_enhancing_2019} which employs a standard multi-layer perceptron DNN.
While ElemNet has been demonstrated for single property prediction, the setting of the present work involves multi-property prediction. \new{ We extended ElemNet to multi-property prediction} \new{(ElemNetMP)} by changing the last layer of the DNN to produce the desired output dimension. Transfer learning with ElemNet was previously reported using transfer learning from a computational to experimental domain wherein both contain the same type of properties. In the present work, our transfer domain has distinct properties from the target prediction domain, 
\new{so} the parameters of the initial layers of the \new{ElemNet model} are initialized by their corresponding values from training in the transfer domain. 

\new{The other baseline models result from the recent incorporation of attention mechanisms in materials property prediction: Roost\cite{goodall_predicting_2020} and CrabNet\cite{wang2020compositionally}. These models were}
also developed for single property prediction, \new{and we extended them to multi-property prediction by changing the last layers of the DNNs to produce the desired output dimension, where we refer to the modified models as RoostMP and CrabNetMP, respectively.} 

We implement \new{RoostMP} \new{and CrabNetMP} using the vector representations of elements provided with their source codes, and while use of these vectors is a type of transfer learning, the transfer is from pre-training with computational formation energy\cite{goodall_predicting_2020,wang2020compositionally}. 
\new{Additionally, we also introduce MP-DOS features to RoostMP and CrabNetMP for transfer learning similar to the HCLMP(T) model, where we call the modified versions RoostMP(T) and CrabNetMP(T) respectively. Specifically, we concatenate the MP-DOS feature vectors with the latent vectors to create a transfer-learned input to the decoders of the RoostMP and CrabNetMP, respectively.}

\subsection*{Settings for multi-target optical absorption prediction}

In the domain of spectral absorption of metal oxides, we consider the prediction space to be the 10-dimensional unitless absorption coefficient ($\alpha \tau$) where the 10 dimensions correspond to equally-spaced ranges of photon energy spanning 1.39 eV to 3.11 eV. The primary dataset was curated from the Materials Experiment and Analysis Database (MEAD)\cite{soedarmadji_tracking_2019} and resulted from high throughput experimental measurement\cite{mitrovic_high-throughput_2015} \new{of complex  metal oxide compositions containing 84628 unique combinations of 39 cation elements. The compositions include 39, 2992, 28612, 50711, and 2274 compositions with 1, 2, 3, 4, and 5 cation elements, respectively. The composition intervals vary among different composition spaces, and some of the high-order composition spaces are from dilute alloy experiments where the data is concentrated around specific points in composition space.} 

Since the standard practice of random test-train splitting of available data does not sufficiently capture the pressing need to make predictions in composition spaces with no available training data, we define each test set by a specific 3-cation composition space.  In the primary settings discussed herein, the combination of 3 cations that defines the prediction composition space is never present in the training set for the respective data instance. Of the 3-cation systems in the dataset, 69 have sufficient data within the 3-cation space as well as the 1 and 2-cation subspaces to be relevant for the settings discussed herein, resulting in 69 unique data instances \new{for which separate models are trained}. For each of these instances, there are up to 36 compositions in the test set, corresponding to the number of unique combinations of 3 elements with 10 at.\% intervals.

\begin{figure}
\includegraphics[width=8.5cm]{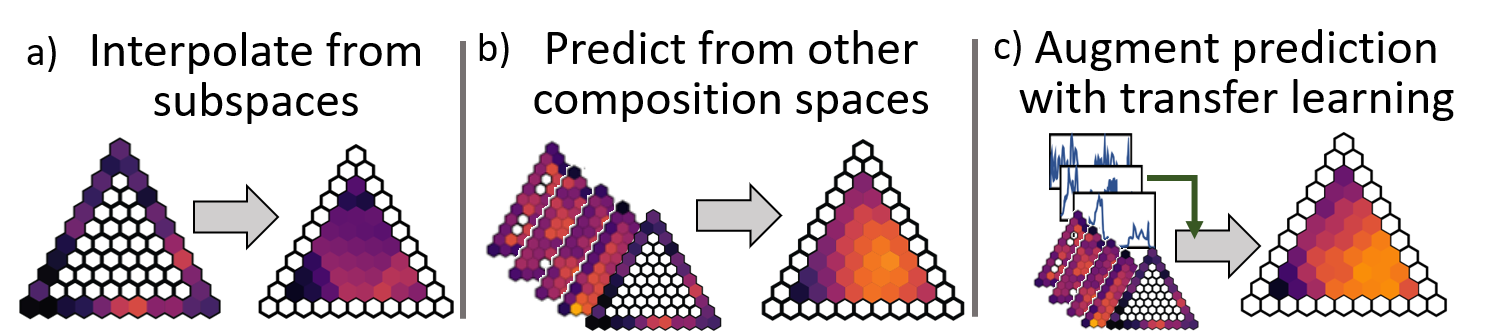}
\caption{
Predicting optical absorption coefficients in new 3-cation composition spaces. Each setting involves prediction of the absorption coefficient at 10 photon energies for each composition in a 3-cation composition space, where the 3 cation elements never coexist in the training data. a) The baseline model LinInterp interpolates from the perimeter of the composition graph. b) ML models learn from other composition spaces to facilitate predictions. c) Predictions can be further augmented via transfer learning, provided the ML model can capture and exploit the chemical knowledge from the transfer domain.
}
\label{fig:settings}
\end{figure}

For the linear interpolation (LinInterp) baseline, only the 30 compositions that lie on the perimeter of the composition graph -- the ternary composition space with 10 atom\% intervals -- are utilized due to the inability to incorporate data beyond the composition graph of the prediction space (Figure \ref{fig:settings}a). The prediction settings for machine learning models expand the training data in the target domain by utilizing any available composition that does not contain the 3 cation elements in the prediction space  (Figure \ref{fig:settings}b). \new{The data available for training and validation} include all 1 and 2-cation compositions, the other 3-cation composition spaces, and the subset of the 4 and 5-cation composition spaces that contain no more than 2 of the 3 cation elements in the prediction space. \new{The validation set is defined as 10\% of these compositions that are first selected from the compositions that contain a pair of the 3 elements that define the test set, followed by random selection used as-needed to reach the 10\% size of the validation set. This definition of validation set is used for selecting the best model observed during the epochs of model training, where neither the training data nor the validation data used for model selection include any composition with all 3 cation elements that define the test set.} The transfer setting employs this same training\new{-validation-test} data in the target domain while additionally utilizing \new{generative} transfer learning from the MP-DOS domain (Figure \ref{fig:settings}c).

\begin{figure}
\includegraphics[width=8.5cm]{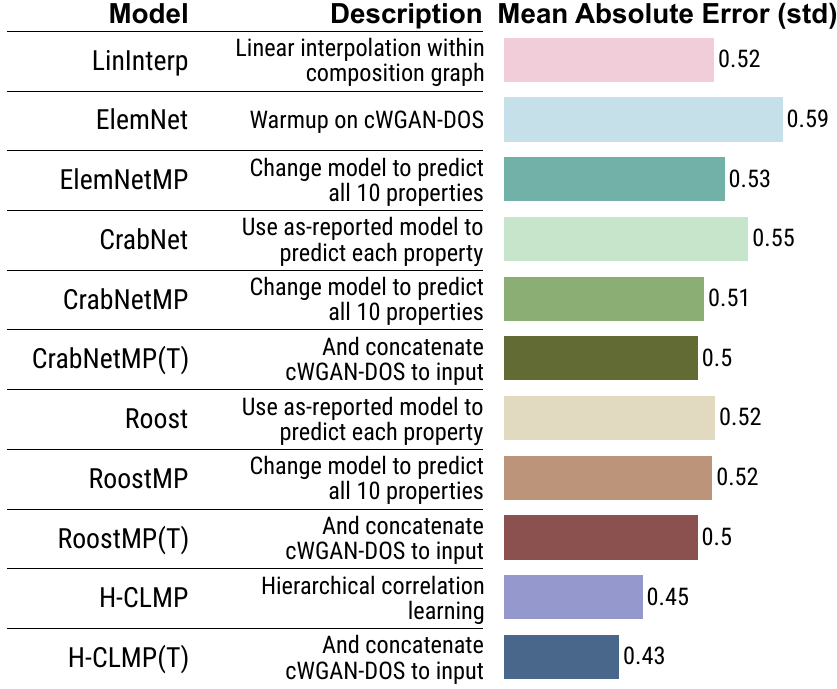}
\caption{Prediction models considered herein that each predict optical absorption in a new 3-cation composition space using only composition as input. \new{LinInterp interpolates each property signal from the compositions subspaces as shown in Figure \ref{fig:settings}. ElemNet is pre-trained on the same MP-DOS data used for generative transfer learning, and ElemNetMP is its extension for multi-property prediction. CrabNet and Roost are attention-based models used as provided by the respective publications, which includes pre-training on computational materials data such as formation energy. Each of these models were modified to enable multi-property prediction with a single model (CrabNetMP and RoostMP) and were additionally used with concatenation of the generative transfer learning. H-CLMP and H-CLMP(T) are the hierarchical correlation models of the present work. For each model, the standardized MAE aggregated over 10 photon energies and 69 data instances is shown as a horizontal bar with numeric label.}}
\label{fig:models}
\end{figure}

For each data instance, the data was normalized to mean-zero and unit-variance, and the mean absolute error (MAE) was calculated in units of the transformation's standard deviation (std), the convention adopted for all reported MAE values in the present work. Composition plots of predicted values use the original $\alpha \tau$ units, obtained via inverse transform of the respective model output. 

The primary comparison for prediction performance is the average MAE over the 69 data instances. To provide context for the MAE \new{of predictions in new composition spaces}, we consider an additional Random setting wherein the test set is defined as 30\% \new{of the data in the union of the 69 3-cation test sets such that prediction is evaluated using compositions that appeared in one of the 69 test sets.} This random-split setting represents a far easier task than the primary prediction setting because the training data contain examples from the test set composition spaces, enabling a model to learn non-linearity within each ternary composition graph directly from the data as opposed to indirectly from other composition spaces and/or domains. Since the data from high throughput experiments has unquantified experimental error, zero MAE an unrealistic goal for prediction models. Given that prediction for a given composition in the Random setting will typically be made only 10 at.\% away from multiple training examples, we expect the resulting prediction error to reflect the experimental error.

\subsection*{Prediction results}

\begin{figure}
\includegraphics[width=8.5cm]{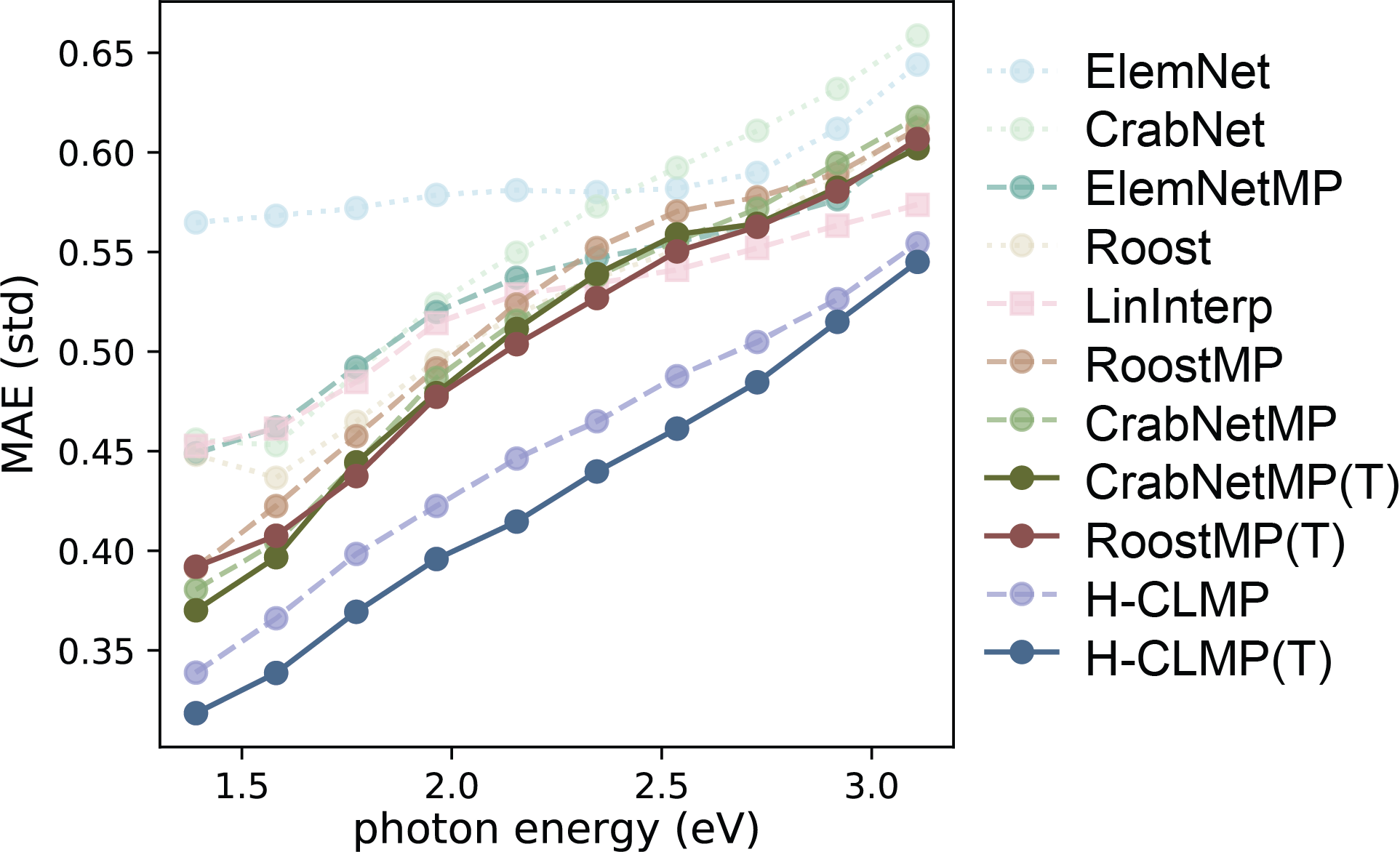}
\caption{The standardized MAE averaged over 69 data instances \new{for each model described in Figure \ref{fig:models}}, plotted as a function of the 10 photon energy ranges of the source data. H-CLMP and H-CLMP(T) outperform all other models. LinInterp operates in the setting of Figure \ref{fig:settings}a by interpolating from the perimeter to the interior of each 3-cation composition triangle. The ML models train using other composition spaces, as shown in Figure \ref{fig:settings}b. The models that employ transfer learning (Figure \ref{fig:settings}c) are the ElemNet models that pre-train on MP-DOS data as well as CrabNetMP(T), RoostMP(T), and H-CLMP(T) that use generated transfer representation of MP-DOS. By transferring knowledge from the MP-DOS domain, generative transfer learning improves performance of each type of model: CrabNetMP, RoostMP, and H-CLMP, respectively. 
}
\label{fig:mae_tern}
\end{figure}

\new{The prediction models are summarized in Figure \ref{fig:models} along with the corresponding average MAE over all photon energies and data instances. The MAE for each photon energy is plotted in Figure \ref{fig:mae_tern} where the 11 models are ordered according to the aggregate MAE shown in Figure \ref{fig:models}. Collectively, these visualizations of model performance reveal several important observations. Comparing the MAE of the LinInterp baseline to the ML models indicates that linear interpolation is rarely outperformed. On one hand, this finding is surprising given the propensity for each of the ML models to learn non-linear relationships. However, one must also consider that LinInterp is predicting only from subspaces as illustrated in Figure \ref{fig:settings}a, which is strongly contrasted by ML model training on the broad range of compositions in the settings of Figure \ref{fig:settings}b-c. A consequence of this localization of the LinInterp model is that the model for each data instance is not applicable for prediction in any new spaces, so the model can not be deployed for the ultimate goal of predicting the spectral absorption of any composition of interest. So while comparison to LinInterp provides evaluation of whether the models accurately predict non-linear composition-property relationships in the test sets, further interpretation of this comparison is limited by the vastly different scope of the prediction tasks in the respective settings. H-CLMP and H-CLMP(T) are the only models that outperform LinInterp at all 10 photon energies.}

\begin{figure*}
\includegraphics[width=14cm]{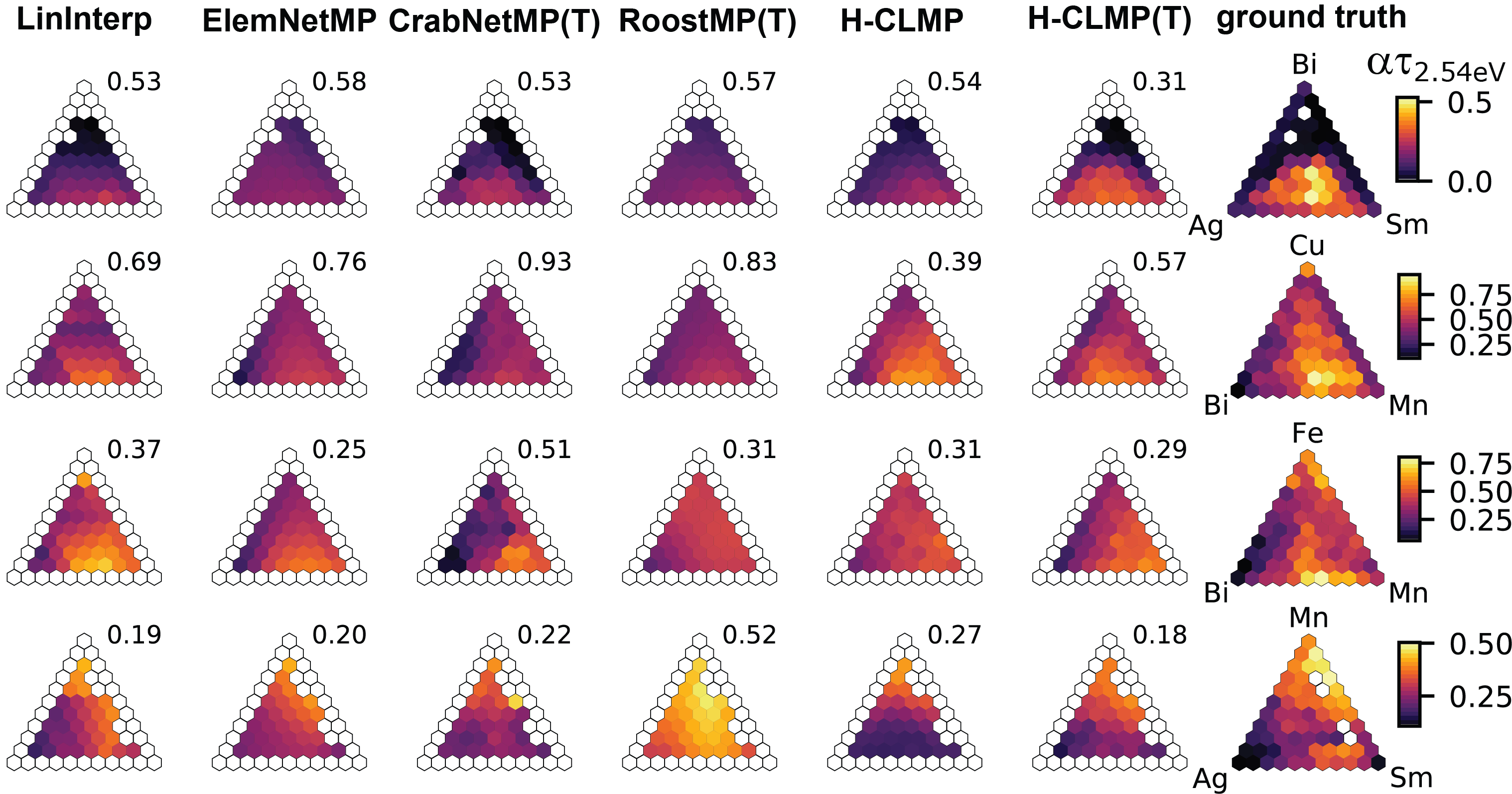}
\caption{The ground truth and predictions for \new{4 of the 69 data instances at a single photon energy, 2.54 eV. The prediction} models are ordered left-to-right by decreasing aggregate MAE \new{as shown in figure \ref{fig:models}}, and each composition plot includes a text label of the standardized MAE for the respective model and photon energy. The cation element labels and the color scale for each photon energy are shown for the ground truth data and apply to all models. For all plots of predicted values, only the compositions in the test set are shown, which excludes the perimeter of the composition graph that was used for training. \new{In the top and bottom rows, some 3-cation compositions are missing in each figure due to their absence in the optical absorption data set. Note that the color scales in this figure correspond to the unitless absorption coefficient, which is not directly comparable to the standardized MAE values of Figures \ref{fig:models}-\ref{fig:mae_rand}.}
}
\label{fig:terns}
\end{figure*}

\new{The generative transfer learning lowers the MAE by creating a more expressive representation of each material, although starting with CrabNetMP, RoostMP, and H-CLMP, the reductions in MAE are a modest 1\%, 3\%, and 5\%, respectively. While these results demonstrate the generative transfer learning concept for enhancing property prediction, the considerable disconnect between computational DOS and experimental absorption spectra limits the ability for models to exploit the knowledge embedded in the generated DOS patterns.} 

\new{The most aggressive baseline models for H-CLMP(T) are our adaptions of the attention-based models from the literature, CrabNetMP(T) and RoostMP(T). These models both perform MPP using the generated DOS transfer learning and provide a practically identical standardized MAE of 0.505. The standardized MAE for H-CLMP(T) is 0.428, a 15\% reduction due to the combination of VAE model alignment and hierarchical correlation learning (see Table \ref{tab:models-compare}). Starting with H-CLMP(T), we performed ablation studies in which i) the hierarchical correlation learning was removed from components (a) and (b) in Figure \ref{fig:H-CLMP} by replacing the multivariate Gaussian and GAT decoders with MLPs, and ii) component (b) was removed to eliminate model alignment during H-CLMP(T) training. The resulting MAEs are 0.454 and 0.455, respectively, which in each case is more than 6\% higher than that of H-CLMP(T). Additionally considering the benefit of the generative transfer learning, these results show that removing any one of the 3 primary advancements of H-CLMP(T), i) hierarchical correlation learning, ii) VAE model alignment, and iii) generative transfer learning, results in an increase in MAE between 5.4\% and 6.3\%. Removing all 3 of these techniques would result in a model similar to RoostMP and CrabNetMP, which provide standardized MAE of 0.519 and 0.511 respectively, both about 19\% larger than that of H-CLMP(T).
}

\new{To further explore the predictions, Figure \ref{fig:terns} shows the ground truth and prediction results for 4 of the data instances at a photon energy of 2.5 eV. These 4 data instances span different qualitative composition-property relationships in the ground truth data as well as different relative performance of the prediction models. The first data instance is Ag-Bi-Sm where the 3-cation ground truth data includes absorption values in a composition region with 10\%-30\% Bi that are higher than any of the absorption values in the subspaces, and the absorption drops precipitously with increasing Bi concentration. While all models predict a general decrease in absorption with increasing Bi concentration, the H-CLMP(T) prediction stands out as the only model that predicts relatively high absorption in the correct 10\%-30\% Bi composition region, which leads to the substantially lower MAE. In the H-CLMP(T) prediction, the Ag$_{0.5}$Bi$_{0.1}$Sm$_{0.4}$ composition is predicted to have the global maximum absorption in the composition graph, higher than any of the compositions in the subspaces. The ground truth shows global maximum at the neighboring composition, Ag$_{0.4}$Bi$_{0.1}$Sm$_{0.5}$, demonstrating excellent qualitative agreement in this surprising discovery of a high-absorption material obtained by mixing 3 cation elements whose individual 1-cation oxides each exhibit much lower absorption. Such predictions demonstrate that the ML models must not simply learn the contributions of each element to the properties, but rather the interactions of the elements that give rise to properties beyond those attainable in composition subspaces.}

\new{The second example is the Bi-Cu-Mn data instance where a local maximum in absorption along the Bi-Mn composition line is further enhanced by addition of 10\% Cu. Since these 3-cation compositions with high absorption are composition neighbors of the subspace compositions, the LinInterp model somewhat captures the composition-property relationships. The LinInterp MAE is only beaten by H-CLMP and H-CLMP(T) where each model correctly predicts high absorption with 10\% Cu and approximately equal amounts of Bi and Mn. While these predictions are qualitatively similar, H-CLMP has substantially lower MAE than H-CLMP(T) in this particular case.}

\new{The final 2 examples are the Bi-Fe-Mn and Ag-Mn-Sm data instances where the composition-property relationships are unremarkable, and as a result there is little headroom for outperforming LinInterp. Collectively, the data instances in Figure \ref{fig:terns} provide perspective on the relative aggregate MAEs shown in Figures \ref{fig:mae_tern} and \ref{fig:models}. In these 4 data instances, the aggregate MAE of H-CLMP(T) is only slightly lower than that of H-CLMP, which is consistent with the aforementioned global decrease in MAE by only 5\% via incorporation of the generative transfer learning. The qualitative composition-property relationships predicted by H-CLMP(T) are consistently correct in Figure \ref{fig:terns}, further confirming that this model is most suitable for guiding materials discovery in new composition spaces.}

\begin{figure}
\includegraphics[width=8.5cm]{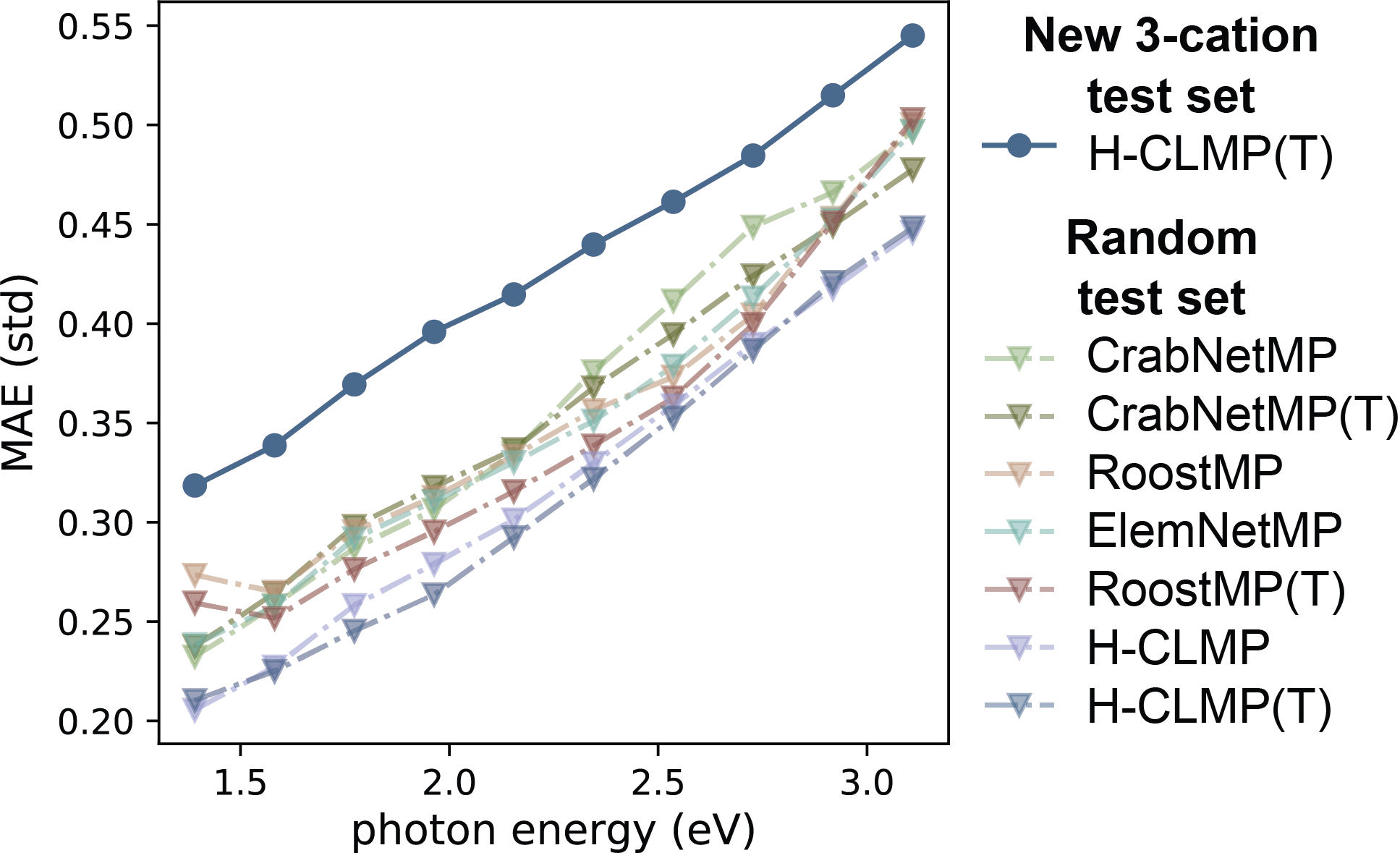}
\caption{\new{The standardized MAE for representative models from Figure \ref{fig:models} deployed in the Random setting where all 69 data instances are combined and the test set contains a random selection of 3-cation compositions. The MAE for H-CLMP(T) from Figure \ref{fig:mae_tern} is shown for comparison.} The Random setting is an easier prediction task that approximates the minimum viable MAE. In this setting, hierarchical correlation learning improves performance over baseline models but generative transfer learning has little impact because the models train on samples in the same composition spaces as the test set.
}
\label{fig:mae_rand}
\end{figure}

\new{To gauge the relative difficulty of the prediction in new 3-cation spaces compared to the common construction of a test set via random selection, the MAE results for the Random setting are shown in Figure \ref{fig:mae_rand}. For most ML models, the MAE is lowered by 29\%-34\% compared to the results of Figure \ref{fig:mae_tern}, with H-CLMP(T) showing a lower 26\% improvement largely because it has the lowest MAE in Figure \ref{fig:mae_tern}. In the Random setting, the MAE data reveal diminishing returns for model performance with increasing model sophistication and with use of generative transfer learning. The relatively similar MAE of the models in the Random setting is a result of the relative ease of this task wherein training data contains compositions that surround the test set compositions. In Figure \ref{fig:mae_tern}, the MAE of each MPP model increases with photon energy,} and given that the results in the Random setting exhibit this same trend, we infer that this trend is due to increasing experimental error with increasing photon energy.

\subsection*{Deployment for Materials Discovery}

\begin{figure}
\includegraphics[width=8.5cm]{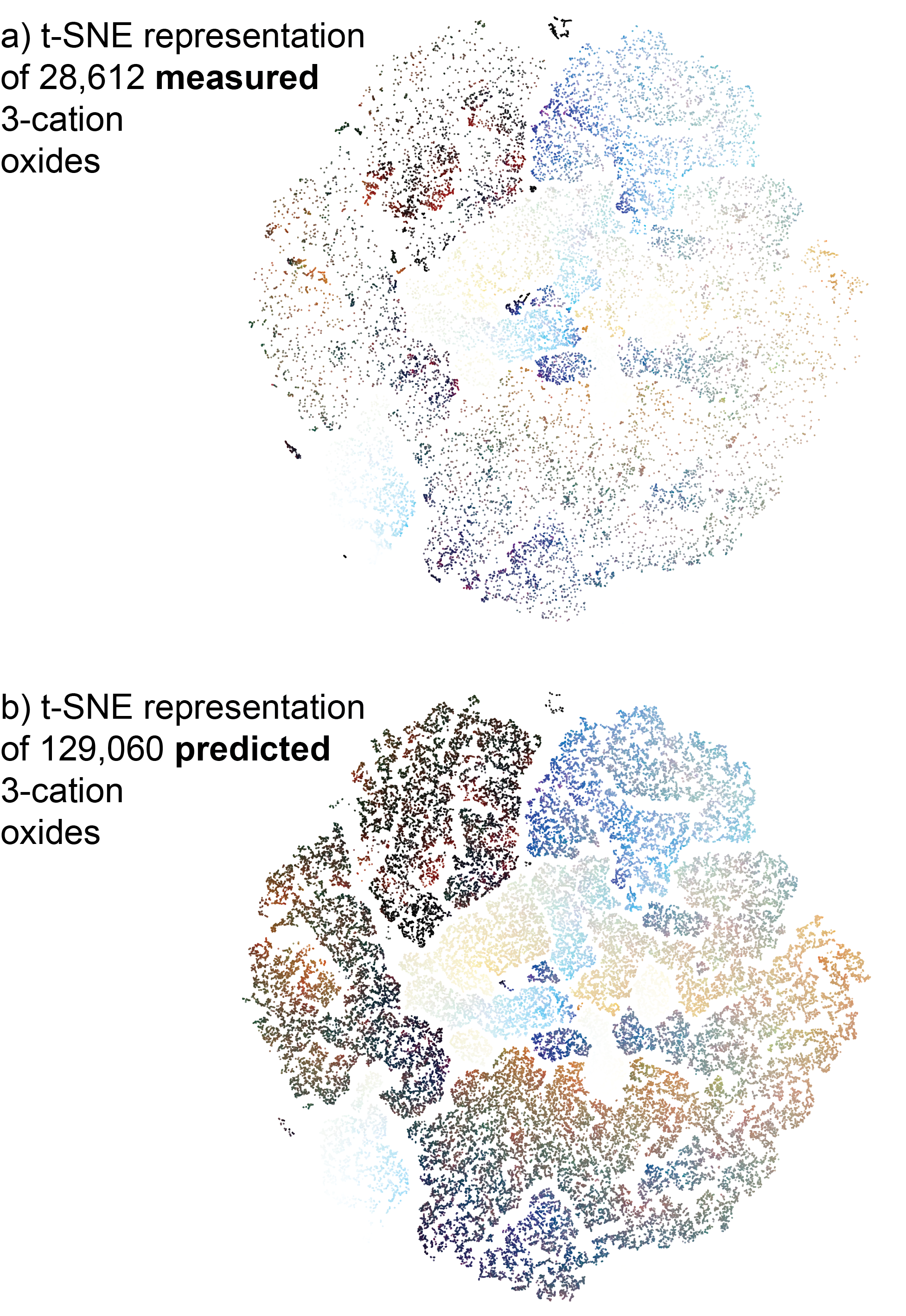}
\caption{\new{Representation of each absorption spectrum of the 3-cation compositions in a) the training set and b) the H-CLMP(T) predictions in new composition spaces, where each composition is plotted at a location determined via 2-dimensional t-SNE mapping of the 10-dimensional absorption spectra, which was trained on the union of the data points in a) and b). Each point is colored according to a quantile-scaled representation of its absorption spectrum. }
}
\label{fig:tsne}
\end{figure}

\new{Given the demonstrated ability of H-CLMP(T) to predict composition-property relationships with good qualitative and reasonable quantitative agreement in the 69 data instance with available ground truth data, we deploy the model to more broadly predict the absorption spectra of complex oxides by considering all possible combinations of the 31 elements that appear in 3-cation compositions in the dataset. Using the same 10\% composition grid in the prediction spaces, this prediction task involves 129060 3-cation metal oxide compositions that are not represented in the 69 data instances described above. To facilitate visualization of this trove of prediction data, we utilize the t-distributed stochastic neighbor embedding (t-SNE) technique\cite{JMLR:v9:vandermaaten08a} to plot each composition's absorption spectra in a 2-dimensional plot, as shown in Figure \ref{fig:tsne}. The t-SNE transformation was trained on the union of all measured and predicted 3-cation compositions, where the measured data includes compositions beyond the 10\% composition grid used to define the 69 data instances described above.}

\new{The range of t-SNE coordinates and colors is similar in the 2 datasets, demonstrating that the H-CLMP(T) predictions do not include absorption spectra that are distinct from the breadth of absorption spectrum in the training set. The H-CLMP(T) model learns the range of absorption spectra that are attainable with complex metal oxide compositions in the training set and predicts the location in that distribution for each new composition. As a result, deployment of the model to guide materials discovery is compatible with identification of new compositions that exhibit a spectral absorption with specific attributes, provided that the training data contains examples of absorption spectra with similar attributes. By using high throughput experiments to provide a breadth of training data that covers a broad range of absorption spectra, the desired absorption spectra attributes can be quite broad.}

\new{For a given materials discovery task, the optical properties are generally not the only properties of interest. Additional materials properties are desirable and may be found with different combinations of 3 cation elements. As a simple example, even if the primary desired property is material's color for use as a dye, additional metrics could include cost, toxicity, etc., and the H-CLMP(T) predictions would enable screening over vast chemistries to identify all candidate materials with the desired optical properties so that they can subsequently be screened for additional properties.}

\new{A common use for spectral absorption data is to identify candidate materials with a desired band gap energy, where down-selection of materials based on their propensity for providing the desired band gap energy leads to subsequent screening for additional properties of interest.\cite{yan_solar_2017} For metal oxide photoanodes, a 2 eV band gap is desirable for solar fuels applications to provide a balance between utilization of the solar spectrum and a sufficiently high possible photovoltage to perform the desired chemistry.\cite{yan_solar_2017} Indeed, much of the present data was collected in search of such materials. While a 10-photon-energy absorption spectrum is insufficient for modelling band gap energy, we can consider related criteria for a 2 eV band gap metal oxide. With increasing photon energy, absorption increases when crossing the band gap energy, motivating criteria based on relatively high absorption above 2 eV compared to the absorption below 2 eV. We thus define a criteria that the average absorption in the 3 absorption spectrum points that span 2.2-2.7 eV be at last 5 times larger than that of the 4 lowest photon energy points that span 1.4-2.0 eV. To avoid identification of transparent materials, the average absorption from 2.2-2.7 eV is also required to be above 0.2.}

\new{Considering 3-cation oxides, 535 of the 28612 compositions in the training set meet these criteria, comprising 47 3-cation composition spaces. In the prediction space, 602 compositions are predicted to meet these criteria, comprising 144 3-cation spaces not in the training set. Naturally, some of these 3-cation spaces will be predicted to meet the criteria because a 2-cation composition in the training data meets the criteria and the model predicts that adding 10\% of an additional cation does not alter the absorption spectrum significantly enough to fail the criteria. The more notable predictions are thus the 3-cation spaces predicted to have a composition meeting the criteria even though none of the 2-cation subspaces in the training data meet the criteria. There are 44 such 3-cation spaces from H-CLMP(T) predictions: Ag-Fe-La, Ce-Fe-Mo, Fe-Mo-Zn, Fe-La-Zn, Mo-Pr-V, Ca-Gd-Pd, Ca-Mg-Pd, Fe-Pd-Zn, Ce-Fe-Zn, Fe-Pd-Yb, Ag-Fe-Pd, Ag-Fe-P, Bi-Cr-Yb, Ce-Pd-Pr, Cu-Fe-Pd, Eu-Fe-Nd, Ca-Pd-Zn, Ca-Co-Pd, Cu-Fe-La, Bi-Cr-Gd, Fe-La-Yb, Fe-K-Yb, Ag-Fe-Mo, Bi-Eu-Mn, Fe-La-Nd, Cu-Fe-Mo, Fe-Mn-Pd, Ca-Ce-Pd, Eu-Fe-Pd, Ag-Fe-Ti, Mn-Mo-Pr, Bi-Cr-Eu, Fe-P-Zn, Ca-Pd-Pr, Bi-Mo-V, Ca-Cu-Pd, Ag-Ca-Pd, Ag-Bi-Cr, Fe-Sr-Yb, Ce-Fe-Pd, Ca-Pd-Sm, Bi-Cr-K, Bi-Cr-Nb, Ca-Ni-Pd. }

\new{Pd appears in 18 of these 3-cation spaces, for which no fundamental explanation is presently available, although inspecting the training data reveals that Pd only exists in 3 3-cation compositions spaces: Bi-Mn-Pd, Bi-Pd-Sm, and Mn-Pd-Sm. All 3 of these spaces contain a composition that meet the criteria, making it understandable that combining Pd with elements similar to those in these 3 training-set composition spaces would result in prediction of absorption spectra that meet the criteria. However, of the 336 Pd-containing 3-cation composition spaces in the prediction set (but not the training set) where none of the subspaces meet the criteria, only the 18 listed above are predicted to contain compositions that meet the criteria, which is only 5.4\% of such 3-cation spaces in the prediction set. The discriminative nature of the predictions are precisely what is needed for guiding materials discovery. This point is further elucidated by considering an additional use case of the 3-cation predictions. Since Pd and Pd-oxide-based catalysts are competent for catalyzing a broad range of chemical transformations, one may seek an oxide with at least 50\% Pd that is transparent across the entire visible spectrum. In the training set, there is only 1 composition that has an absorption value below 0.22 for all 10 photon energies: Bi$_{0.05}$Pd$_{0.7}$Sm$_{0.25}$. In the prediction set, there is 1 additional such composition, Co$_{0.1}$Mo$_{0.4}$Pd$_{0.5}$. How the model ascertains that a Pd-rich composition with Co and Mn will have an absorption spectrum similar to its only training example, a mixture with Bi and Sm, cannot be ascertained at this time, yet these use cases demonstrate that H-CLMP(T) is making nontrivial predictions involving its learned interactions among elements and cWGAN-generated DOS. These use cases also demonstrate how such a prediction model can serve as a rapid down-selection mechanism for guiding materials discovery efforts, in the present case for materials with specific spectral absorption attributes.}

\section*{Discussion}

We have proposed a novel machine learning framework for multi-property prediction that integrates state-of-the-art machine learning techniques - latent embedding learning, correlation learning, generative transfer learning, and attention models. Latent embedding learning is a recent technique for matching labels and features in the latent space. Pioneer studies~\cite{yu2014large,chen2012feature,bhatia2015sparse} made low-rank assumptions on labels and features, transforming labels to label embeddings by dimensionality reduction techniques. Recent methods that are based on deep learning are believed to implicitly encode the label correlations in the embedding space.\cite{sundar2020out} Our model learns probabilistic feature and label subspaces under a variational autoencoder (VAE) framework. The latent spaces are aligned with KL-divergence, and the sampling process enforces smoothness. The label correlations can be taken into account through three different strategies, namely zeroth-order, first-order, and higher-order. The zeroth-order strategy converts the problem into multiple, independent single-label prediction problems and ignores the correlations among labels.\cite{boutell2004learning} The first-order strategy considers the pairwise correlations between labels, as has been demonstrated in other fields.\cite{huang2015learning,huang2017multi,nan2018local} Higher-order approaches mine the relationship among many labels.\cite{yapp2020comparison,tsoumakas2009mining} Our hierarchical correlation learning employs both the first-order, pairwise label correlations via the multivariate Gaussian and the higher-order label correlations via the attention model.

The attention
mechanism is the crux behind many state-of-the-art
sequence-to-sequence models used in machine translation and language processing  \cite{gonzalez2021} and it has recently shown good results on multi-label classification.\cite{lanchantin2019neural} While the attention mechanism has also been recently adopted to perform learning of relationships among elements in material property prediction, \cite{wang2020compositionally,goodall_predicting_2020} our model \new{additionally} uses the attention mechanism to perform learning of relationships among multiple material properties by acting on the output of the multivariate Gaussian model as opposed to the composition itself.

The DNN of the ElemNet model has the inherent flexibility to learn correlations among multiple properties, although the model's lack of an explicit representation of correlation increases the reliance on training data to learn these relationships.
As a result, with sufficient training data, models such as DNNs may become competitive with H-CLMP, \new{although such training data is often limited in materials science}.

Many types of transfer learning have been developed in a variety of domains.\cite{zhuang_comprehensive_2021} The implementations of transfer learning in the ElemNet\cite{jha_enhancing_2019} and Roost\cite{goodall_predicting_2020} models are based on a warm-up strategy in which the model architecture for property prediction is pre-trained using data from the transfer domain. This approach requires the transfer domain to conform to the model architecture of the target domain, whereas generative transfer learning, as employed by H-CLMP(T), alleviates this restriction and enables use of transfer learning models tailored to the transfer domain. Furthermore, multiple transfer domains can be employed to aggregate knowledge in parallel, as opposed to the serial utilization of transfer domain data in the warm-up strategy. The physical sciences are ripe with use cases wherein a variety of datasets offer different perspectives related to the desired prediction task, and H-CLMP(T) provides a framework that can exploit multiple knowledge sources, especially when intuition of expert scientists guides development of models that extract the desired knowledge from the transfer domain(s). 

Collectively, the results highlight the importance of integrating multiple AI techniques to provide predictive models in new composition spaces. \new{The integration of AI techniques in the present work build upon the foundation of our prior work} \new{developing general multi-label classification and and multi-target regression approaches that were initially 
motivated by ecology applications. These prior works demonstrated the utility of multivariate Gaussian used for pairwise correlation learning,\cite{chen2018end} model alignment with a VAE during training,\cite{kong2020deep,bai2020disentangled} and high-order correlation learning via an attention graph neural network.\cite{zhao2021hot}} 

Our careful crafting of model architecture is particularly motivated by the need to make predictions in new composition spaces. For the much easier task of predicting properties of compositions similar to those in the training data, a more straightforward model may be sufficient. Indeed, Figure \ref{fig:mae_rand} shows that in the Random setting, the MAE from \new{all ML models} are all similar. The broader lesson for development of machine learning models in the physical sciences is that realistic settings with aggressive prediction tasks are necessary to demonstrate the predictive power of advanced algorithms. The collection of 69 data instances for multi-target regression presented herein, which each involve prediction into new composition spaces \new{that are} completely hidden from training, is well suited as a benchmark machine learning dataset for the community.

\section*{Conclusion}
We present the H-CLMP(T) framework to seamlessly integrate complementary AI techniques and tackle challenging prediction tasks in the physical sciences. The use of each individual AI method for materials property prediction constitutes a seminal demonstration of these state-of-the-art techniques in materials science. Most notably, we introduce model architectures that learn element and property relationships at multiple scales, as well as a broadly-applicable generative transfer learning approach 
to augment prediction in the target domain. Collectively, these techniques integrate multiple knowledge sources to predict properties of new composition spaces. \new{Based on a set of 31 elements with 3-cation training data, predicted absorption spectra are provided for in new 3-cation composition spaces to guide materials discovery efforts.} The prediction of properties in high-order composition spaces is a grand challenge of AI-assisted materials discovery that is well addressed by combining hierarchical correlation learning with generative transfer learning.

\section*{Methods}

\subsection{Dataset curation}
The primary dataset was curated from the Materials Experiment and Analysis Database (MEAD)\cite{soedarmadji_tracking_2019} and resulted from high throughput optical spectroscopy measurements described previously.\cite{mitrovic_high-throughput_2015} Briefly, metal oxide samples were deposited by inkjet printing of 1 to \new{5} solutions each containing an elemental precursor. The deposits were then dried and calcined in air or synthetic air (20\% O$_2$ in N$_2$) for at least 10 minutes at a temperature of at least 300 $^{\circ}$C. Typical annealing conditions were 400 $^{\circ}$C for 6 hours in air, although some variability in processing conditions was permitted for the present work to enable assembly of a dataset spanning many composition spaces. All samples were confirmed to have been processed in only air or synthetic air, conditions known to produce metal oxide samples for all elemental precursors considered herein. Transmission and reflection ultraviolet-visible spectroscopy was used measure optical absorption using data processing methods described previously.\cite{suram_high_2016} 
The unitless absorption coefficient was calculated as the negative of the natural logarithm of $T/(1-R)$, where $T$ and $R$ are the fractional transmission and reflection signals, respectively. \new{Measurement error or anti-reflection properties of the metal oxides can cause negative absorption coefficients when using this model, which are unphysical with respect to the interpretation of the measured absorption coefficient as the optical attenuation coefficient in the bulk of the material, but in the case of anti-reflective coatings negative values can result from the amount of light captured by the transmission and reflection measurements in the presence of the materials exceeding that captured in the absence, which occurs to the imperfect light collection of the high throughput spectrometers. Negative values occur in 1.8\% of the values in the entire dataset but only 0.7\% of the values in the union of all 69 test sets, suggesting that their impact on the results and conclusions is negligible}. 

The molar loading of each sample was designed to be similar although not explicitly measured, prompting our approximation of the molar optical absorption coefficient as being proportional to the unitless absorption coefficient, where consideration of the proportionality constant is irrelevant for the present work that uses zero-mean, unit-variance scaled data. The unitless absorption coefficient was averaged in 10 equally-sized and equally-space ranges of photon energies whose center photon energies span 1.39 eV to 3.11 eV. 
The oxygen concentrations were unknown and excluded from the composition representation. For cation compositions with multiple available measurements meeting the above criteria, the mean optical absorption coefficient over composition duplicates was used to represent the composition. Note that experimental error in both composition and optical absorption measurements, as well as any influences from variability in film morphology or processing conditions, were not considered. Collectively, these imperfections in the data result in a minimal reasonable MAE for model prediction, which as noted above is characterized using the Random-Global setting.

\new{The data instances were identified in the source dataset by considering only compositions with up to 3 cation elements that lie on a 10 atom\% composition grid. For a given 3-cation space, there are 36 such compositions that contain all 3 cation elements, and 30 compositions in the subspaces, i.e. compositions that contain 1 or 2 of the 3 cation elements. To identify data instances with approximately equal amounts of subspace training data and 3-cation test data, each data instance is required to have at least 25 of the possible 30 subspace compositions and at least 26 of the 36 possible 3-cation compositions. From the set of 39 elements, the 3-cation compositions that meet these criteria include 69 unique combinations from a set of 29 elements: }
Ag, Bi, Ca, Ce, Co, Cr, Cu, Er, Eu, Fe, Gd, K, La, Mg, Mn, Mo, Nb, Nd, Ni, P, Pd, Pr, Sm, Sr, Ti, V, W, Yb, Zn. \new{The dataset additionally includes cation elements Ba, Ga, Hf, In, Pb, Rb, Sc, Sn, Tb, and Zr}
that only appear in training data. \new{Of these, only Sc and Zr appear in 3-cation compositions on a 10 atom\% grid with the above set of 29 cation elements, whereas the other 8 elements primarily appear in dilute alloy compositions. As a result, when considering the setting for predicting in all possible composition spaces, only the 31 elements that appear in the dataset as 3-cation compositions on a 10\% grid these elements are used to define the prediction set. The resulting prediction set is the 4426 3-cation combinations of the 31 elements that are not one of the 69 data instances.}

The transfer data was obtained by searching version 2020.3.13  of the Materials Project database for entries containing only oxygen and any of the 39 elements listed above. Furthermore, only entries containing total DOS data over a range including -8 eV to 8 eV were considered, resulting in 41322 entries. For present purposes, only the composition of non-oxygen elements was used to represent each entry, and the DOS for each entry was resampled to a common energy grid from -8 eV to 8 eV with 0.1 eV intervals. \new{The DOS dataset was randomly split into 80\% for training, 10\% for validation, and 10\% for testing. The test set is only to confirm that the cWGAN is predictive for DOS, as demonstrated by its standardized MAE of 0.43.}

\subsection{H-CLMP Model}

Our model's overall architecture is depicted in Figure \ref{fig:H-CLMP}. 
For the cWGAN in component (c), we use a 3-layer multilayer perceptron (MLP) as the generator, and the 3 layers have 256, 512, and 256 neurons, respectively. We use an additional MLP as its critic, where the critic has 3 layers, and the 3 layers have 256, 256 and 128 neurons, respectively. The cWGAN
uses the earth mover's distance as its loss function and is pre-trained on the density of state (DOS) from the material project (MP) dataset. In component (d), we then transfer the knowledge learned from the material project (MP)  dataset to help with learning a more accurate predictor for the smaller experimental datasets. The 
DOS 
representation generator $\bar{G}$ of component (d) 
is fixed after training component (c). There are many possibilities for configuring $\bar{G}$, and here we 
perform sampling over the random input $Z$ of the cWGAN. With 100 random samples of $Z$, $G$ produces 100 generated DOS arrays, whose mean generated DOS is defined as the output of $\bar{G}$. This average generated DOS is used as an additional input to components
(a) and (b). 
\new{The feature encoder in component (a) is an attention graph neural network, which is adopted from the Roost model.\cite{goodall_predicting_2020}
}
\new{The label encoder in component (b) is a 2-layer MLP with 256 neurons in each layer.}

\new{To align the latent representations in components (a) and (b), the final part of each encoder is 
a multivariate Gaussian with a diagonal covariance matrix of size 128 $\times$ 128. Note that the since this covariance matrix is diagonal, this multivariate Gaussian is not part of the heirarchical correlation learning but rather a mechanism to provide smooth, aligned outputs of the respective encoders. 
}

\new{The decoders in components (a) and (b) have parallel construction and commence with pairwise property correlation learning via a multivariate Gaussian model in each component with a shared covariance matrix. Each multivariate Gaussian is } 
parameterized by mean vectors $\mu$ of length $10$ and \new{shared} covariance matrix $\Sigma$ of size $10\times 10$. The multi-property embeddings \new{of size $10\times 512$} are generated by sampling from their corresponding multivariate Gaussian latent spaces.
\new{Higher-order property correlation learning proceeds via an } attention graph neural network, whose description can be found in prior literature. \cite{goodall_predicting_2020,wang2020compositionally,gonzalez2021,lanchantin2019neural} We use 5 attention layers, namely the message-passing operations are executed 5 times. Each attention layer also includes a element-wise feed-forward MLP which has 2 layers of 128 neurons each. 

\new{The loss function for simultaneously training components (a) and (b) is the MAE combined with the KL divergence loss described above, with relative weights of 1 and 0.1, respectively.}
H-CLMP(T) is implemented with the Pytorch deep learning framework. The model was trained with the Adam optimizer \new{with \new{an initial} learning rate 0.0005 \new{that was halved after 20 epochs}, a batch size of 128, and  \new{training total of} 40 epochs} in an end-to-end fashion in a machine with NVIDIA V100 16GB GPUs. \new{Model selection was based on the minimum MAE of the validation set.
}

\new{Ablation studies for H-CLMP(T) proceeded as follows: For removing VAE alignment, component (b) was not used, resulting in component (a) being trained independently with no KL divergence term in the training loss function. For removing hierarchical correlation learning, the decoder in each component (a) and (b) was replaced by a standard 3-layer MLP decoder with 256 neurons in each layer and with model parameters shared between components (a) and (b).}

\subsection{Baseline models}
\new{Linear interpolation was performed using available composition on the 10\% composition grid in the subspaces of the 3-cation test set. Each test set composition and each subspace composition for training was represented as a normalized 3-component composition: $(a, b, 1-a-b)$. Transformation to 2-D Euclidean coordinates proceeded as $x = 1-a-b/2$ and $y = b*3^{1/2}/2$. This is the same transformation used for the ternary composition plots. The 2-D linear interpolation was then performed using the griddata function in numpy version 1.18.1.}

\new{ElemNet, Roost, and CrabNet were developed for single property prediction, and we extended them to multi-property prediction by changing the last layers of the DNNs to produce the desired output dimension, where we refer to the modified models as ElemNetMP, RoostMP, and CrabNetMP, respectively. ElemNet 
\new{was pretrained} 
with the MP-DOS dataset,
\new{and then all but the last MLP layer was transferred to continue training in the optical absorption domain.}
We implemented RoostMP and CrabNetMP using the vector representations of elements provided with the source code, and while use of these vectors constitutes transfer learning, the transfer is from computational formation energy\cite{goodall_predicting_2020,wang2020compositionally}. 
Additionally, we also introduced MP-DOS features to RoostMP and CrabNetMP for transfer learning similar to the HCLMP(T) model, where we refer to the modified versions as RoostMP(T) and CrabNetMP(T), respectively. Specifically, we concatenated 
\new{each generated DOS vector to the respective latent vector from the model's encoder, creating a latent representation with MP-DOS transfer learning that served as the input to the model's decoder.
All baseline models were trained for 40 epochs with model selection based on the minimum MAE of the validation set.
}
}

\subsection{Visualization of prediction data}

\new{We used t-distributed stochastic neighbor embedding (t-SNE) to reduce the 10-dimensional absorption spectra to 2 components for visualization. The scikit-learn t-SNE implementation was applied to the union of measured ternary composition sets used for model training and the unmeasured (predicted) ternary composition sets, with parameters $perplexity$=15 and $n\_iter$=5000. Points in the 2-D scatterplot were colored by assigning the mean alpha values in the ranges 1.39-1.77, 1.96-2.35, 2.53-3.11 eV, to red, green, and blue (RGB) color channels respectively. A quantile transformation was applied to rescale the RGB values to a uniform distribution and utilize a wider range of the color space. Both t-SNE dimensionality reduction and quantile scaling were applied to the union of the measured and predicted 3-cation data.}

\begin{acknowledgments}
This work was funded by the U.S. Department of Energy, Office of
Science, Office of Basic Energy Sciences, under Award DE-SC0020383 (data curation, design of multi-property prediction setting and transfer setting, model evaluation) and by the Toyota Research Institute through the Accelerated Materials Design and Discovery program (development of machine learning models).
The authors thank Santosh K. Suram for assistance with curation of the dataset.
\end{acknowledgments}

\section*{Data Availability}
\new{The data that support the findings of this study are available at https://data.caltech.edu/records/1878 doi: 10.22002/D1.1878. 
The source code  and additional data for H-CLMP are available at https://www.cs.cornell.edu/gomes/udiscoverit/?tag=materials.}


\nocite{*}
\section*{References}
\bibliography{references}

\end{document}